\documentclass[runningheads]{llncs}
\usepackage{graphicx}
\usepackage[T1]{fontenc}
\usepackage[utf8]{inputenc}
\usepackage{xcolor}
\usepackage{makecell}
\usepackage{cite}

\def\FINAL{}

\begin{document}
\title{Predicting Multiple ICD-10 Codes from Brazilian-Portuguese Clinical Notes}
%
%

\ifdefined\FINAL
	\author{Arthur D. Reys\inst{1,2} \and
	Danilo Silva\inst{1} \and
	Daniel Severo\inst{2} \and
	Saulo Pedro\inst{2}\and
	Marcia M. de Sousa e Sá\inst{3}\and
	Guilherme A. C. Salgado\inst{2}}
	\authorrunning{A. D. Reys et al.}
	\institute{Federal University of Santa Catarina. Florianópolis, Brazil \\
	\email{danilo.silva@ufsc.br} \and
	3778 Healthcare. Belo Horizonte, Brazil\\
	\email{\{arthur.reys, severo, saulo.pedro, guilherme\}@3778.care} \\
	\url{https://3778.care/} \and
	Syrian-Lebanese Hospital. São Paulo, Brazil \\
	\email{marcia.sa@hsl.org.br}}
\fi

\maketitle

\begin{abstract}
ICD coding from electronic clinical records is a manual, time-consuming and expensive process. Code assignment is, however, an important task for billing purposes and database organization. While many works have studied the problem of automated ICD coding from free text using machine learning techniques, most use records in the English language, especially from the MIMIC-III public dataset. This work presents results for a dataset with Brazilian Portuguese clinical notes. We develop and optimize a Logistic Regression model, a Convolutional Neural Network (CNN), a Gated Recurrent Unit Neural Network and a CNN with Attention (CNN-Att) for prediction of diagnosis ICD codes. We also report our results for the MIMIC-III dataset, which outperform previous work among models of the same families, as well as the state of the art. Compared to MIMIC-III, the Brazilian Portuguese dataset contains far fewer words per document, when only discharge summaries are used. We experiment concatenating additional documents available in this dataset, achieving a great boost in performance. The CNN-Att model achieves the best results on both datasets, with micro-averaged F1 score of 0.537 on MIMIC-III and 0.485 on our dataset with additional documents.
	
	
\end{abstract}

\section{Introduction}

Throughout the stay of a patient in a hospital, a series of documents are written about their situation, including symptoms, clinical evolution, diagnoses and medical history. After the release of a patient, medical coders analyze their documentation and assign to that stay a list of codes based on the International Classification of Diseases (ICD), a standard system maintained by the World Health Organization~\cite{organizationInternationalClassificationDiseases1978,organizationICD10InternationalStatistical2004}. Those codes identify a variety of clinical information, which is useful for billing purposes, health plan communication and organizing databases for research and statistical analysis~\cite{jensenMiningElectronicHealth2012a}.

Currently the ICD coding process is manually performed by specifically trained coders. The granularity of the coding system makes differences between similar codes very subtle. Moreover, much of the information in clinical records comes in unstructured free text and the language used is specific to the medical field, containing abbreviations, ambiguous terms and typos. Together, those factors make manual coding an expensive, time consuming and error-prone task.

The development of machine learning models over free text from Electronic Health Records (EHR) for automated ICD coding has been discussed for over two decades~\cite{larkeyAutomaticAssignmentICD91995}. Recently, models based on natural language processing techniques using advanced neural networks have shown relevant performance improvements~\cite{mullenbachExplainablePredictionMedical2018,liICDCodingClinical2019}. However, most of these works involve English-language data. To the best of our knowledge, only~\cite{delimaHierarchicalApproachAutomatic1998,ferraoUsingStructuredEHR2013a,oleynikAutomated2017,santosUsingDeepConvolutional2018,duarteDeepNeuralModels2018} have considered a Portuguese-language dataset. Except for~\cite{duarteDeepNeuralModels2018}, which focuses on a different task of coding the causes of death from death certificates, and~\cite{oleynikAutomated2017}, which aims at predicting groups of oncology ICD codes from pathology reports, all others use small datasets to predict a limited set of ICD codes. Also, none provide comparisons with accessible datasets.

In this work, we consider the problem of automatically assigning multiple diagnostic ICD codes to a patient stay based on Brazilian Portuguese free-text clinical notes, considering all available codes. Specifically, we develop and compare Logistic Regression (LR), Convolutional Neural Network (CNN), Recurrent Neural Network (RNN) and CNN-based attention models with optimized hyperparameters. We present a case study based on data from Syrian-Lebanese Hospital, a Brazilian hospital in São Paulo, where we intend to deploy our best performing model in order to support the ICD tagging process. Additionally, we provide results for the publicly available English-language dataset MIMIC-III (Medical Information Mart for Intensive Care)~\cite{johnsonMIMICIIIClinical2016,johnsonMIMICIIIFreelyAccessible2016}, where we outperform previous work among models of the same families and the current state of the art.\footnote{Code for MIMIC-III is available at https://github.com/3778/icd-prediction-mimic.}

\section{Background}

\subsection{Previous Work}

In the ICD coding task, researchers often have to decide which codes will be the target of the study. While some works consider all types of ICD codes~\cite{xuMultimodalMachineLearning2019}, others use a limited amount of ICD codes~\cite{santosUsingDeepConvolutional2018} or limit the scope to Diagnoses ICD codes~\cite{liAutomatedICD9Coding2019a}. This is done mainly because of differences in datasets and the large class imbalance observed in the majority of them. As free text inputs for this specific task, most works use discharge summaries, as they condense information about a patient stay in a single document~\cite{mullenbachExplainablePredictionMedical2018}. However,~\cite{duarteDeepNeuralModels2018} and~\cite{xuMultimodalMachineLearning2019} have experimented using additional documents.

The structure of the ICD system is used to develop a hierarchical approach to assist predictions in~\cite{baumelMultiLabelClassificationPatient2017} and~\cite{perotteDiagnosisCodeAssignment2014}. A method based on ICD co-occurrence is proposed in~\cite{subotinMethodModelingCooccurrence2016}. In~\cite{crammerAutomaticCodeAssignment2007}, overlaps between ICD descriptions and words in documents compose a rule-based method. More prominently, works use machine learning models such as SVM (Support Vector Machine)~\cite{perotteDiagnosisCodeAssignment2014}, Naive Bayes~\cite{pakhomovAutomatingAssignmentDiagnosis2006,medoriMachineLearningFeatures2010a} and kNN (k-Nearest Neighbors)~\cite{ruchEpisodesCareDiagnosis2008}.

Convolutional Neural Networks (CNN) have been widely used in the literature, achieving good results in the ICD coding task~\cite{liAutomatedICD9Coding2019a,liConvolutionalNeuralNetworks2017,mullenbachExplainablePredictionMedical2018}. The advantage of this architecture over more traditional machine learning models (such as LR and SVM) is its capability of capturing local contextual features~\cite{liAutomatedICD9Coding2019a}. Recurrent Neural Networks have also been used due to their ability to associate information in longer contexts than CNNs~\cite{huangEmpiricalEvaluationDeep2019,[ayyarTaggingPatientNotes2017],baumelMultiLabelClassificationPatient2017}. In particular, LSTM (Long Short-term Memory) and GRU (Gated Recurrent Unit) recurrent networks capture information within a large contextual window. These approaches have achieved improvements over older machine learning models, as free text usually have high complexity and their comprehension rely on local and global semantic relations between terms and sequences.

In addition to neural networks, innovative models include ensemble of different architectures~\cite{xieNeuralArchitectureAutomated2018,xuMultimodalMachineLearning2019} and \textit{per-label} attention mechanisms~\cite{liICDCodingClinical2019,mullenbachExplainablePredictionMedical2018}. Per-label attention consists of weighing a base representation of documents differently for each ICD code. In the specific task of this work, including only Diagnoses ICD codes,~\cite{mullenbachExplainablePredictionMedical2018} appears to hold the current state of the art.

Due to the limited availability of public EHRs, most works focus on MIMIC \cite{johnsonMIMICIIIFreelyAccessible2016}, a freely accessible dataset in English language. Works aimed at EHRs in Portuguese are rare and use different private data sources~\cite{delimaHierarchicalApproachAutomatic1998,ferraoUsingStructuredEHR2013a,oleynikAutomated2017,santosUsingDeepConvolutional2018,duarteDeepNeuralModels2018}. Among these,~\cite{delimaHierarchicalApproachAutomatic1998} shows how an hierarchical approach can improve results in the ICD coding task. An approach based solely on structured data is presented in~\cite{ferraoUsingStructuredEHR2013a}. In~\cite{santosUsingDeepConvolutional2018}, a CNN with self-taught GloVe embeddings is presented to predict a small set of possible codes from free text, while a cost-sensitive learning approach is implemented to overcome class imbalance. These works use relatively small datasets and focus on few codes. In turn,~\cite{oleynikAutomated2017} and~\cite{duarteDeepNeuralModels2018} use large collections of data. In~\cite{oleynikAutomated2017}, SVM is used to predict groups of topographical and morphological oncology ICD codes from pathology reports, in a one-\textit{versus}-all approach. Finally,~\cite{duarteDeepNeuralModels2018} uses a recurrent neural network with attention to predict ICD codes corresponding to death causes from death certificates and related documents. However, oncology ICD groups and death causes ICD codes still comprise smaller sets than diagnostic codes, while pathology reports and death certificates have significant structural, semantic and lexical differences from clinical notes such as discharge summaries.

\subsection{Feature Extraction}

Training a computational model over free text requires some kind of feature extraction method. Among different methods, some encode whole documents into vectors, without regard to the order of the words. This is called a Bag-of-Words (BoW) representation, with the most popular approach being TF-IDF (Term Frequency -- Inverse Document Frequency)~\cite{saltonTermweightingApproachesAutomatic1988}. Others generate latent vector representations of words, as Word2Vec~\cite{mikolovEfficientEstimationWord2013}, GloVE~\cite{penningtonGloveGlobalVectors2014} and FastText~\cite{bojanowskiEnrichingWordVectors2017}, allowing documents to be represented as a sequence of word vectors. More enhanced methods at word level include ELMo~\cite{petersDeepContextualizedWord2018} and BERT~\cite{devlinBERTPretrainingDeep2019}. In these methods, the same word can be mapped to different vectors, depending on their surrounding context. Other methods include character-level and paragraph-level representations~\cite{zhangCharacterlevelConvolutionalNetworks2016,leDistributedRepresentationsSentences2014}. 

In this work we use TF-IDF features for Logistic Regression and Word2Vec for the neural networks. Hence, a detailed description of these methods is given.

\subsubsection{Term Frequency -- Inverse Document Frequency}

TF-IDF~\cite{saltonTermweightingApproachesAutomatic1988} aims to reflect the importance of a word in a document, given a \textit{corpus}. Based on a BoW model, a document is converted into a multi-hot encoding of words contained in it, based on vocabulary constructed from the \textit{corpus}. Then, words are given weights based on their importance for each document. The importance of a word increases proportionally to the number of times it appears in that document (term frequency) and inversely proportional to the total of documents that contain it (inverse document frequency).

\subsubsection{Word2Vec}

Word2Vec~\cite{mikolovEfficientEstimationWord2013} is a representation model that takes into account order and context of words in documents. The inputs are tokenized texts and the model builds a vocabulary associating words to correspondent fixed-dimension vectors. Tokens in the \textit{corpus} are projected into a multi-dimensional space, allowing identification of interdependent relations between different terms, through cosine similarity.

Word2Vec is composed of a single hidden layer neural network. Two methods can be applied in the embedding training: CBoW (Continuous Bag-of-Words) and Skip-gram. In CBoW, a word is predicted from a limited amount of words that precede and succeed it. The context words are converted into BoW features, losing local ordering information between them. In Skip-gram, the task is to predict, from a given word, a limited amount of words around it. In this case, the order of the context words influences the network projection, as nearby words receive higher weights. The latent word representations resulting from Word2Vec training can be loaded as an embedding layer in neural network based models. An embedding layer is a mapping of discrete input variables (e.g. tokens representing words) to corresponding vector representations.

\section{Datasets and Preprocessing}

\subsection{MIMIC-III Dataset}

The MIMIC-III dataset---the third revision of MIMIC, v1.4---is a publicly accessible English-language dataset that includes numerous tables relative to patients in Beth Israel Deaconess Medical Center, in the United States~\cite{johnsonMIMICIIIFreelyAccessible2016}. Each admission of a patient to the hospital is associated to several documents, as well as to an ordered list of ICD codes, using the Diagnoses ICD-9-CM (where CM stands for Clinical Modification) coding system, at the most specific level (i.e. subcategories).

As the majority of related works, only free text discharge summaries were selected, totaling 52722 hospital admissions from 41127 unique patients. We found a total of 6918 unique ICD codes associated with these documents.

We perform light preprocessing on the input texts, removing date/hour patterns, special characters and applying lowercase. The same data split as in~\cite{mullenbachExplainablePredictionMedical2018} was used, consisting of 47719 samples in the training set, 1631 in the validation set, and 3372 in the test set. In this split, no patient is listed in more than one subset.

\subsection{HSL Dataset}
\label{subsec:hsl_dataset}

The HSL dataset contains de-identified documents linked to patients from the Syrian-Lebanese Hospital (HSL). Collected between 2016 and 2018, texts are written in Brazilian Portuguese. The dataset includes different types of documents in free text. Each document has a hospital admission ID from which different documents can be linked. We removed all admissions that did not have a linked discharge summary, totaling 77005 admissions from 51298 unique patients. Each admission has a list of ICD codes tagged by professional medical coders using Diagnoses ICD-10 codes at the most specific level (i.e. subcategories). We found 5360 unique codes in the dataset.

Initially, we selected only discharge summaries (S), with each admission containing a single document. This set is referenced as HSL-S. However, after further analysis, we decided to include additional free text documents which were numerously available, in particular: clinical developments (E) and anamnesis/physical exams (A). Unlike discharge summaries, a wide range of types E and A documents are attached to each admission, from none to several texts. Table~\ref{tab:dataset_document_types}
\begin{table}[t]
	\centering
	\caption{Statistics of document types in MIMIC-III and HSL datasets.}
	\begin{tabular}{c|c|c|c|c}
		        Dataset          & Unique patients & Admissions & Total documents & Avg. words per sample$^{\mathrm{a}}$ \\ \hline\hline
		MIMIC-III$^{\mathrm{b}}$ &      41127      &   52722    &      52722      &                 1327.5                  \\ \hline
		         HSL-S           &      51298      &   77005    &      77005      &                  94.6                   \\ \hline
		         HSL-E           &      50899      &   76159    &     919713      &                 1483.0                  \\ \hline
		         HSL-A           &      42153      &   59249    &      63423      &                  155.4                  \\ \hline
		        HSL-SEA          &      51298      &   77005    &     1060141     &                 1730.4                  \\ \hline
		\multicolumn{5}{l}{$^{\mathrm{a}}$Concatenation of all documents corresponding to the same admission.}              \\
		\multicolumn{5}{l}{$^{\mathrm{b}}$Only discharge summaries.}
	\end{tabular}
	\label{tab:dataset_document_types}
\end{table}
shows the total of documents per type and the unique admissions and patients linked to these documents. These additional documents were concatenated to type S documents with the same admission ID, to form the input text for each sample. This better reproduces the human coding process that takes place at Syrian-Lebanese Hospital, where coders observe all documents of an admission to determine the correspondent ICD codes. We refer to the dataset with concatenated types S, E and A documents as HSL-SEA.

Text preprocessing is done in the same way as in MIMIC-III. We split data ensuring no patient was present in more than one subset, totaling 69309 samples in the training set; 2313 in the validation set; and 5383 in the test set.

\subsection{Comparison Between Datasets}

\begin{figure}[t]
	\centering
	\includegraphics[width=0.7\textwidth]{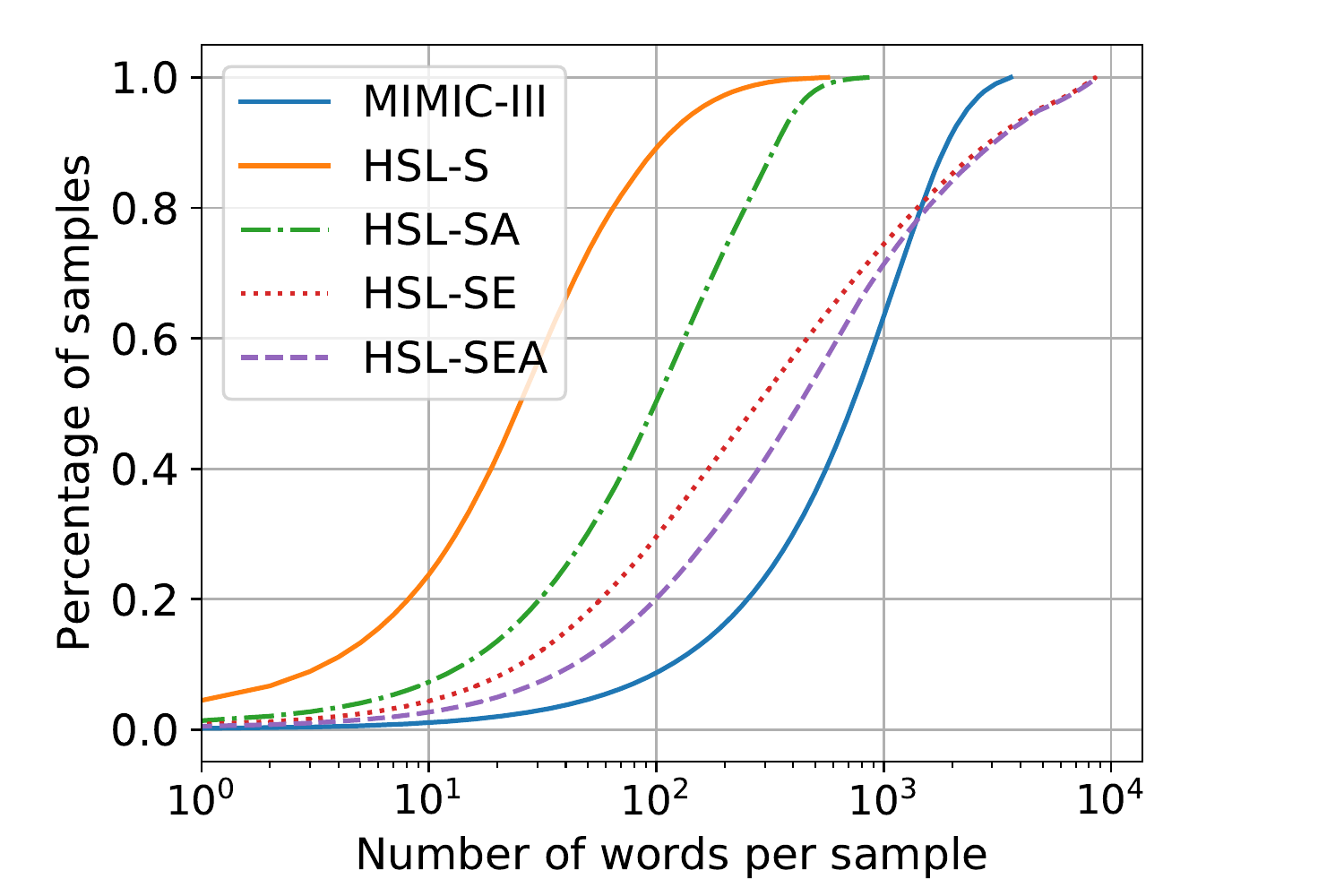}
	\caption{Word count per sample cumulative distribution on all datasets.}
	\label{fig:datasets_word_cumsum}
\end{figure}

Besides language, the presented datasets have some relevant differences. Fig.~\ref{fig:datasets_word_cumsum}
shows the cumulative distribution of word count per sample in all datasets, after text preprocessing. Table~\ref{tab:dataset_document_types} presents differences in the number of documents selected for each dataset. As also shown in Table~\ref{tab:dataset_document_types}, MIMIC-III discharge summaries have a much larger average of words per sample than HSL-S. The concatenation of S, E and A documents to HSL-SEA result in an average closer to MIMIC-III. From these statistics, we can assert that MIMIC-III discharge summaries contain, objectively, far more data than HSL-S, while having a closer average and distribution to HSL-SEA. Also, by looking at random samples, we noticed more detailed and well written texts in MIMIC-III. 

The ICD coding systems adopted by the datasets are also different. While MIMIC-III uses a Clinical Modification of ICD-9, HSL uses the newer ICD-10.

\begin{figure}[t]
	\centering
	\includegraphics[width=0.7\textwidth]{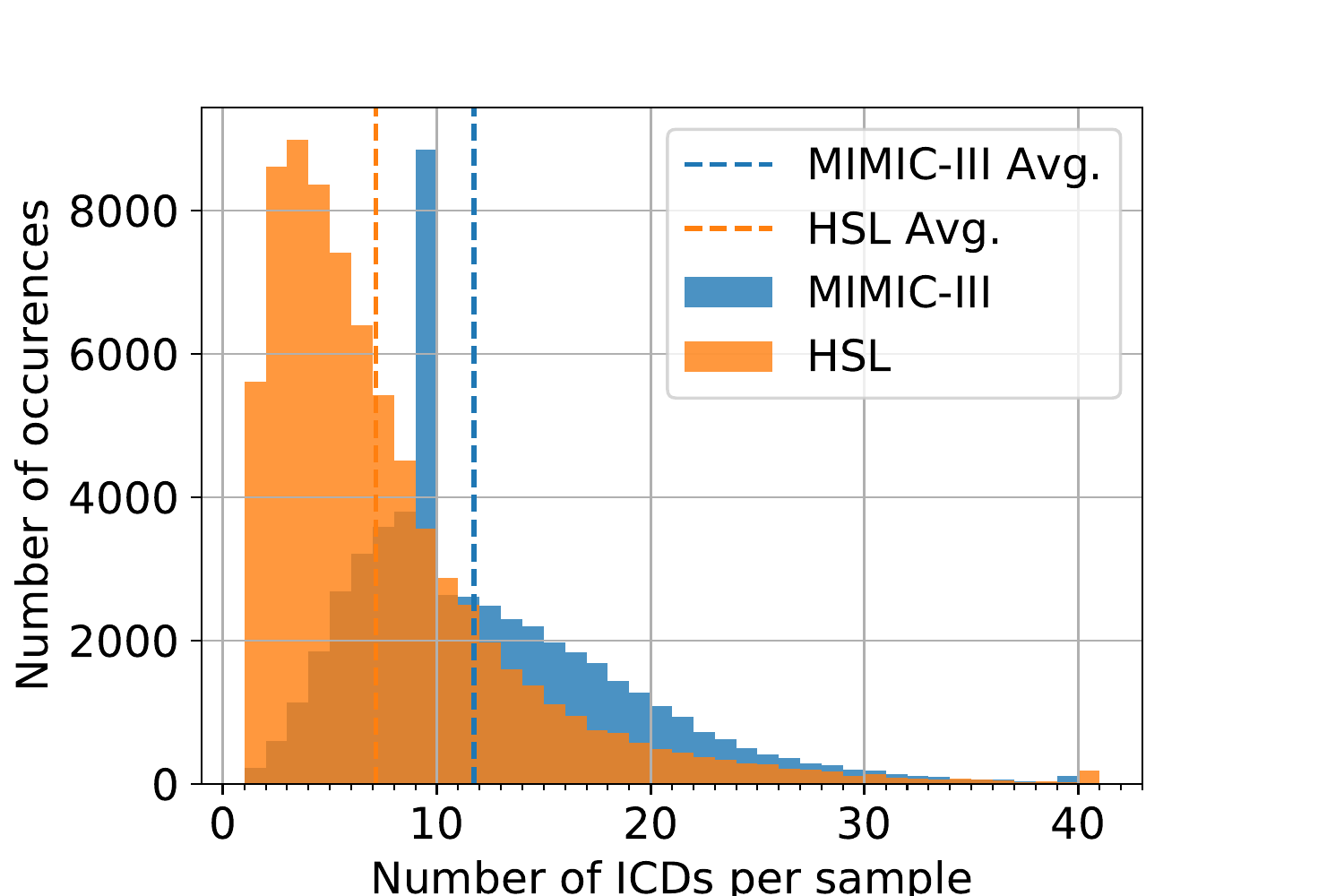}
	\caption{ICD count per sample histograms on MIMIC-III and HSL.}
	\label{fig:datasets_icd_hist}
\end{figure}

Fig.~\ref{fig:datasets_icd_hist} shows histograms of the number of ICD codes per sample for both datasets.
While maximum, minimum and standard deviation are similar,
the average number of ICD codes per sample is lower in HSL. Note that the classes are extremely imbalanced in both datasets, as some examples show in Table~\ref{tab:datasets_icd_dist}. We also note that 4.47\% and 4.48\% of the ICD codes contained in the test sets are not present in the training sets, respectively, in MIMIC-III and HSL.

%
\begin{table}[t]
	\centering
	\caption{Percentage of samples tagged with the 1st, 10th, 100th and 1000th most frequent ICD codes on MIMIC-III and HSL.}
	\begin{tabular}{c|c|c}
		Dataset & MIMIC-III &   HSL   \\ \hline\hline
		  1st   &  38.02\%  & 34.37\% \\ \hline
		 10th   &  11.67\%  & 10.71\% \\ \hline
		 100th  &  2.23\%   & 1.26\%  \\ \hline
		1000th  &  0.15\%   & 0.06\%
	\end{tabular}
	\label{tab:datasets_icd_dist}
\end{table}

\section{Methods}

In this section we present the evaluation metrics and models used in this work. 

\subsection{Evaluation Metrics}

We used popular metrics for multi-label tasks, namely, F1, precision and recall, all micro-averaged over different classes~\cite{liAutomatedICD9Coding2019a}. Micro-averaging presents a more representative result considering the large and imbalanced sets of classes, and is indeed used in most works that do not limit the number of ICD codes.

Micro-averaged precision and recall are defined, respectively, as
\begin{equation}
	P_{\textrm{micro}}= \frac{\sum_{c=1}^{C} \sum_{n=1}^{N} y_{n}^{c} \hat{y}_{n}^{c}}{\sum_{c=1}^{C} \sum_{n=1}^{N} \hat{y}_{n}^{c}},
	\qquad 
	R_{\textrm{micro}}= \frac{\sum_{c=1}^{C} \sum_{n=1}^{N} y_{n}^{c} \hat{y}_{n}^{c}}{\sum_{c=1}^{C} \sum_{n=1}^{N} y_{n}^{c}},
\end{equation}
where $C$ is the number of classes, $N$ is the number of samples and $y_n$ and $\hat{y}_n$ are, respectively, true and predicted vectors with $C$ binary entries, each indicative of a class $c$ in a sample $n$. The F1 score (higher is better) is defined as the harmonic mean between precision and recall.

Each model (with the exception of the Constant model) outputs, per sample, a vector with $C$ real-valued entries between 0 and 1 corresponding to the confidence of prediction for each class. In particular, if the model was trained on only $C'<C$ classes, we assign $\hat{y}_{n}^{c}=0$, $\forall n$, for the remaining classes $c$ not seen by the model. In order to compute the above metrics, we analyze a range of thresholds to binarize outputs, selecting the best one for each model based on F1 in the validation set.

\subsection{Models}

The models developed in this work are described below. We used the Keras\footnote{http://keras.io/} framework with Tensorflow\footnote{https://tensorflow.org/} backend in all implementations. Models were trained to a maximum of 10 epochs. Instead of applying Early Stopping, after each epoch we computed F1 in the validation set. When training was over, we restored weights corresponding to the epoch with the best result. For our study we used an AWS EC2 virtual machine with 8 vCPUs and a NVIDIA T4 GPU.

\subsubsection{Constant (Top-$k$)}

The objective of this baseline model is to determine whether the performance of real models is greater than that of an implementation which does not use the input texts. 

The Constant model predicts a constant list of $k$ ICD codes for all samples. The ICD codes selected are the $k$ most occurring in the training set. The parameter $k$ was optimized to obtain the best F1 in the validation sets, resulting in $k=15$ for MIMIC-III and $k=8$ for HSL.

\subsubsection{Logistic Regression}

In the LR model, we convert the multi-label problem into a set of binary classification problems, one for each class. The inputs of the LR model are TF-IDF features computed over each dataset.

TF-IDF was implemented using Scikit-learn\footnote{https://scikit-learn.org/}. Stopwords were removed from the preprocessed texts, using default Portuguese and English stopwords from Natural Language Toolkit\footnote{https://www.nltk.org/}. Maximum vocabulary size was fixed to the 20000 most frequent words.

The hyperparameters of the LR were optimized via Grid Search considering different optimizers, learning rates from 0.0001 to 0.1 in multiples of 10 and L2 regularizer parameters from 0 (no regularization) to 10, also in multiples of 10. The final model uses Adam optimizer with learning rate 0.1 and all other optimizer parameters set to default values. No regularization is performed. Each training epoch took 50 seconds for MIMIC-III and 60 seconds for HSL-S and HSL-SEA.

\subsubsection{Convolutional Neural Network}
\label{subsubsec:cnn_model}

The CNN implemented in this work consists of an embedding layer loaded with Word2Vec word vectors, followed by a single one-dimensional convolutional layer and Batch Normalization~\cite{ioffeBatchNormalizationAccelerating2015}. On the output, a Global Average Pooling operation precedes a fully connected layer with as many units as the number of classes for each dataset. It is based on the implementation of~\cite{mullenbachExplainablePredictionMedical2018}, but with some modifications: our tests showed that removing Dropout, adding Batch Normalization and increasing kernel size from 4 to 10 improved results, as well as performing Global Average Pooling instead of Global Max Pooling. The layers and respective parameters are shown in Table~\ref{tab:models_params}. We used Adam optimizer with learning rate 0.001 for MIMIC-III and 0.003 for HSL-SEA.

\begin{table}[t]
	\centering
	\caption{Architectures and parameters for the neural network models.}
	\begin{tabular}{c|c|c}
		                        CNN                         &          GRU           &                       CNN-Att                       \\ \hline\hline
		                       Input                        &         Input          &                        Input                        \\ \hline
		               Embedding (size 300)                 &  Embedding (size 300)  &                Embedding (size 300)                 \\ \hline
		\makecell[c]{Conv1D \\(500 filters, kernel 10, tanh)} & \makecell[c]{GRU \\(500 units, tanh)}  & \makecell[c]{Conv1D \\(500 filters, kernel 10, tanh)} \\ \hline
		                Batch Normalization                 &  Batch Normalization   &                 Batch Normalization                 \\ \hline
		              GlobalAveragePooling1D                & GlobalAveragePooling1D &                      Attention                      \\ \hline
		                 Output (sigmoid)                   &    Output (sigmoid)    &                  Output (sigmoid)
	\end{tabular}
	\label{tab:models_params}
\end{table}

Given that the CNN model involves Batch Normalization, it is mandatory for the inputs to have fixed sizes~\cite{ioffeBatchNormalizationAccelerating2015}. However, samples have a large variation in number of words, as shown in Figure \ref{fig:datasets_word_cumsum}. To ensure a fixed-length input, texts with fewer words than needed were padded with padding tokens by the end, while texts with more than the maximum of words had their end truncated. The padding token points to a null vector in the embedding layer. Observing the distribution of text sizes among the datasets, the fixed-length of the inputs was set to: 2000, for the MIMIC-III dataset; 300, for HSL-S; and 4000, for HSL-SEA. 

Word2Vec vectors were trained using Gensim\footnote{https://radimrehurek.com/gensim/}. The embeddings were self-trained due to the specificity of the Brazilian Portuguese clinical language, containing medical terms, abbreviations and acronyms~\cite{santosUsingDeepConvolutional2018}. Words appearing in less than 10 samples were not considered. We experimented vector lengths between 100 and 600, CBoW and Skip-gram implementations, and whether stopwords should be removed. These parameters were optimized for the HSL-S dataset, resulting in vectors with length 300, Skip-gram training algorithm and stopwords not being removed. 

Each epoch took 310 seconds when training for MIMIC-III and 820 seconds for HSL-SEA.

\subsubsection{Recurrent Neural Network}
\label{subsubsec:rnn_model}

The RNN model consists of an embedding layer loaded with Word2Vec word vectors, followed by a GRU layer. Then, Batch Normalization and Global Average Pooling are performed. In the output we define a fully connected layer with as many units as the number of classes for each dataset. As in the CNN model, the samples were processed to fit in a fixed-length input, in this case to allow faster training on the GPU. In this work, we used GRU layers for their better results over traditional RNNs, while keeping a simpler architecture (and being more quickly trainable) than LSTMs~\cite{chungEmpiricalEvaluationGated2014}.

Three base architectures were tested: the first one is such as shown in Table~\ref{tab:models_params}; the second has an extra GRU layer; the last has a bidirectional GRU instead of a common GRU. The first architecture yielded the best results, so each parameter was then individually optimized from this base architecture.

Among the optimized parameters are: optimizer; learning rates from 4e-4 to 1e-2 in steps of 1e-4; masking of padding tokens, to avoid their influence on model predictions; sample weighting inversely proportional to the number of true ICD codes; fine-tuning of the embedding layer; and Pooling methods. Adam optimizer with 8e-4 learning rate resulted in improvements in F1, so as fine-tuning the embedding layer. Average Pooling proved to be greatly superior than Max Pooling. The final architecture is shown in Table~\ref{tab:models_params}. This model is referred simply as GRU in the next sections.

The GRU model uses the same Word2Vec vectors trained for the CNNs. Training times per epoch were 268 seconds for MIMIC-III and 785 seconds for HSL-SEA.

\subsubsection{Convolutional Neural Network with Attention}

The CNN model with Attention (CNN-Att) is based on the current state of the art CAML (Convolutional Attention for Multi-Label Classification) \cite{mullenbachExplainablePredictionMedical2018}, with some modifications. The model is also similar to our conventional CNN model, with the only difference that the Global Pooling is replaced by a \textit{per-label} attention mechanism (which computes a separate context vector for each label as a weighted average of the input sequence) and each fully-connected sigmoid output unit takes as input only its corresponding context vector. The attention operation is a scaled dot-product \cite{vaswaniAttentionAllYou2017} and uses a separate trainable target vector for each label (see \cite{mullenbachExplainablePredictionMedical2018} for details). 

Compared to the original CAML model, we removed Dropout from the embedding layer, which in our initial experiments did not seem to improve the performance, and added Batch Normalization after the convolutional layer, since it typically allows for a faster convergence of training. We increased the number of filters in the convolutional layer from 50 to 500. These modifications improved metrics in our tests. Also, to allow faster convergence, we scheduled the learning rate to start at 0.001 in the first two epochs, and only then decrease to 0.0001. Table \ref{tab:models_params} presents the architecture and parameters used in the CNN-Att. 

Following our other neural network models, we used Word2Vec word embeddings, and the samples were processed to fit a fixed-length input (see the CNN model subsection). Training the CNN-Att took 1600 seconds per epoch for MIMIC-III and 3700 seconds per epoch for HSL-SEA.

\section{Results and Discussion}

We trained our models using MIMIC-III and HSL datasets. This section shows achieved results and comparisons, as well as experiments regarding additional documents in HSL.

\subsection{MIMIC-III Results}

Table~\ref{tab:mimic_thresh_res} shows the results obtained for all models on the MIMIC-III test set. As baselines for comparison, we also present results from other works in the literature. 

The Constant model achieves very poor results, as expected. Our LR with optimized hyperparameters greatly outperforms similar LR~\cite{mullenbachExplainablePredictionMedical2018} and SVM~\cite{liAutomatedICD9Coding2019a} linear models, presented as baselines in these works. This suggests that these models were underfitting due to lack of hyperparameter optimization; indeed, we noticed that the LR from~\cite{mullenbachExplainablePredictionMedical2018} used a default L2 regularization parameter of $1$, while we adopted no L2 regularization. The F1 achieved by our LR is comparable to CNN implementations with Word2Vec features found in~\cite{liAutomatedICD9Coding2019a} and~\cite{mullenbachExplainablePredictionMedical2018}, while our CNN shows an improvement over these models. The GRU returns significant improvements over all previous models, as well as over a similar model presented in~\cite{mullenbachExplainablePredictionMedical2018}. Finally, the CNN-Att outperforms all other models, including the original CAML~\cite{mullenbachExplainablePredictionMedical2018}.

\begin{table}[t]
	\centering
	\caption{Performance of different models on MIMIC-III dataset. Entries with no citation brackets correspond to our models.}
	\begin{tabular}{c|c|c|c|c}
		                         Model                           & Threshold &       F1       & Precision & Recall \\ \hline\hline
		                        Constant                         &     -     &     0.192      &   0.188   & 0.196  \\ \hline
		  LR~\cite{mullenbachExplainablePredictionMedical2018}   &     -     &     0.242      &     -     &   -    \\
		       flat-SVM~\cite{liAutomatedICD9Coding2019a}        &     -     &     0.253      &   0.635   & 0.158  \\
		                           LR                            &   0.19    &     0.406      &   0.425   & 0.388  \\ \hline
		 CNN~\cite{mullenbachExplainablePredictionMedical2018}   &     -     &     0.402      &     -     &   -    \\
		         CNN~\cite{liAutomatedICD9Coding2019a}           &     -     &     0.399      &   0.440   & 0.366  \\
		                          CNN                            &   0.30    &     0.423      &   0.467   & 0.387  \\ \hline
		Bi-GRU~\cite{mullenbachExplainablePredictionMedical2018} &     -     &     0.393      &     -     &   -    \\
		                          GRU                            &   0.32    &     0.468      &   0.543   & 0.412  \\ \hline
		 CAML~\cite{mullenbachExplainablePredictionMedical2018}  &     -     &     0.524      &     -     &   -    \\
		                        CNN-Att                          &   0.28    & \textbf{0.537} &   0.590   & 0.492
	\end{tabular}
	\label{tab:mimic_thresh_res}
\end{table}

\subsection{HSL Results}

For the HSL dataset, we first selected only discharge summaries (HSL-S), to allow a more direct comparison with MIMIC-III, which uses only this type of document. As HSL-S and MIMIC-III are very different datasets, we did not expect identical results. Even so, when training the LR model, the results we obtained were much lower than expected, namely, an F1 of 0.316, which is 20\% below that of MIMIC-III.

These results, as well as the fact that HSL-S has a considerably lower average of words per sample than MIMIC-III, lead our study to experiment with other documents available in HSL. We trained the LR model on different combinations of concatenated documents: types S and A; types S and E; and types S, A and E (refer to Section \ref{subsec:hsl_dataset} for an explanation of each document type). Table~\ref{tab:hsl_thresh_lr} presents metrics computed over the validation set. Clearly, adding documents to discharge summaries---thus increasing average words per sample---shows improvements in metrics, with a large increase in F1 when using HSL-SEA.

\begin{table}[t]
	\centering
	\caption{Validation metrics of LR model trained over HSL considering different concatenated document types.}
	\begin{tabular}{c|c|c|c|c}
		Documents  & Threshold &       F1       & Precision & Recall \\ \hline\hline
		    S      &   0.26    &     0.316      &   0.320   & 0.312  \\ \hline
		 S and A   &   0.25    &     0.347      &   0.359   & 0.336  \\ \hline
		 S and E   &   0.27    &     0.357      &   0.382   & 0.336  \\ \hline
		S, E and A &   0.25    & \textbf{0.367} &   0.390   & 0.346
	\end{tabular}
	\label{tab:hsl_thresh_lr}
\end{table}

Considering the outcomes of these experiments, we then trained all models on HSL-SEA. As CNN and RNN are sensitive to the order of concatenation of documents, we experimented orders S-A-E and S-E-A. We adopted the latter one, as it achieved slightly better results. Compared to HSL-S, we achieved consistently better results when using HSL-SEA, for all models. 
Metrics on the HSL-SEA test set are shown in Table~\ref{tab:hsl_thresh_res_safe}. Once more, the CNN is slightly superior than the LR, while the GRU model shows improvements over both of those models. The CNN-Att model presents again the best results, significantly ahead of all other models.

\begin{table}[t]
	\centering
	\caption{Performance of different models for HSL-SEA dataset.}
	\begin{tabular}{c|c|c|c|c}
		 Model   & Threshold &       F1       & Precision & Recall \\ \hline\hline
		Constant &     -     &     0.203      &   0.183   & 0.228  \\ \hline
		   LR    &   0.25    &     0.368      &   0.400   & 0.340  \\ \hline
		  CNN    &   0.26    &     0.374      &   0.386   & 0.363  \\ \hline
		  GRU    &   0.29    &     0.441      &   0.508   & 0.390  \\ \hline
		CNN-Att  &   0.29    & \textbf{0.485} &   0.543   & 0.438
	\end{tabular}
	\label{tab:hsl_thresh_res_safe}
\end{table}

Note that each model on HSL-SEA achieves a performance comparable to (up to about 10\% below) that same model on MIMIC-III. This is evidence that HSL-SEA has comparable quality to MIMIC-III discharge summaries for ICD code prediction.

\section{Conclusion}

This work presented a study on automated ICD coding from free text, using four learning models trained on two datasets. For MIMIC-III, we reproduced and improved results of similar models in the literature, outperforming the state of the art on the prediction of diagnosis codes from discharge summaries. Results show that using a CNN with per-label attention outperforms conventional CNN, GRU and LR models, attaining a Micro-F1 of 0.537.

For the HSL dataset, we observed that using only discharge summaries was insufficient to achieve results similar to MIMIC-III. Besides the different coding system, word count statistics and detail levels in documents may explain the loss in performance. After concatenating additional documents found in HSL, we observed a significant improvement. Again, the best performance was achieved by our optimized CNN-Att model, with a Micro-F1 of 0.485.

We believe our best model trained on HSL could be suited to assist medical coders using clinical records in Brazilian Portuguese, allowing for gains in efficiency and a decrease in errors in the manual ICD tagging process. We are working towards the deployment of a pilot trial to test the usefulness of the model and better understand its limitations in a practical setting.

\ifdefined\FINAL

\section*{Acknowledgments}

The authors would like to thank Ricardo Giglio, Dr. Mauro Cardoso, Dr. Flávio Amaro, and Marcio Gregory for helpful discussions, as well as 3778 Healthcare and Syrian-Lebanese Hospital for their support of this research.

\fi

\bibliographystyle{splncs04}
\bibliography{zotero_references_2}

\end{document}